\documentclass[letterpaper]{article}
\usepackage{aaai}
\nocopyright
\usepackage[utf8]{inputenc}

\usepackage{times}
\usepackage{url}
\usepackage{colonequals}

\usepackage{amsmath,amsfonts,amssymb}
\usepackage{mathtools,bm}

\usepackage{graphicx}
\usepackage{algorithmic}
\usepackage{algorithm}
\usepackage{enumerate}

\usepackage{multirow}

\usepackage{tikz}
\usetikzlibrary{arrows,automata,positioning}

\usepackage{mdframed}

\title{Uncertainty Aware AI ML: Why and How}
\author{Lance Kaplan \\ ARL, USA \And Federico Cerutti\\ Cardiff University, UK \And Murat Sensoy\\ Ozyegin University, Turkey \AND Alun Preece \\ Cardiff University, UK \And Paul Sullivan \\ Intelpoint Inc. USA}

\begin{document}

\maketitle
\begin{abstract}
This paper argues the need for research to realize uncertainty-aware artificial intelligence and machine learning (AI\&ML) systems for decision support by describing a number of motivating scenarios. Furthermore, the paper defines uncertainty-awareness and lays out the challenges along with surveying some promising research directions. A theoretical demonstration illustrates how two emerging uncertainty-aware ML and AI technologies could be integrated and be of value for a route planning operation.
\end{abstract}

\section{Introduction}

The recent successes of machine learning (ML) \cite{hinton2006fast,krizhevsky2012imagenet} have sparked new interest in artificial intelligence (AI)\&ML systems \cite{LeCun:2015}.  For the most part, the emphasis is on improving the accuracy of machine learning.  However, it is known that ML techniques can only interpolate from the data that trained them.  They cannot extrapolate knowledge about a test sample that is far different than the training data. Likewise, AI systems are crafted to reason over a set of concepts manually built by domain experts or learned by examples though a constrained inductive process~\cite{russell2016artificial}.  Thus, AI systems are unable to extend to high-order reasoning to `connect the dots’ the way a human can.  However, humans are flawed by imperfect logical reasoning exacerbated by a number of cognitive biases, e.g. confirmation bias, group-think, loss aversion, etc.

It is crucial for AI\&ML systems to work seamlessly with human agents to complement the strengths of each while abating their flaws.  However, most AI\&ML systems infer without any internal evaluation of the quality of the inference relative to how they were trained. This means that the system can drive a human agent, or worse yet an automated controller, to a flawed decision, e.g., driverless automobile accidents \cite{dickson2018}. In this paper, we argue that in many different scenarios, it is important for AI\&ML systems to evaluate uncertainty in the inference based upon the similarity of the situation under test to the training data and the amount of relevant training data. Furthermore, we argue that it is critical for the AI\&ML system to communicate its uncertainty along with an explanation of the source of the uncertainty to the human decision maker. We refer to such AI\&ML systems as \textit{uncertainty-aware}.  We believe that uncertainty-awareness enables the human to contextually process the inferences for the AI\&ML system along with its confidence to either directly decide or to ponder further. Further ponderance includes the human using his/her creative higher-order reasoning abilities to extrapolate new inferences or seek out more information to formulate the decision.

Within the `Anticipatory Situational Understanding for Coalitions' project, we view an automated decision aid as a layered system where lower-level ML methods operating on raw data feed classification results to a higher-level AI probabilistic decision logic~\cite{harborne2018}.  Both the lower and higher layers of this AI\&ML system should be uncertainty-aware.  This paper will review some of our recent work to enable uncertainty-awareness at both layers inspired by subjective logic~\cite{Josang2016-SL}.  The higher layer is a subjective Bayesian network that is able to infer subjective opinions, i.e., uncertain probabilities, of latent variables given uncertain conditional probabilities.  The lower layer is a neural network that is specially trained so that the output can be interpreted as a subjective opinion. Then, we show how the two layers could form an uncertainty-aware AI\&ML system for a route planning scenario, and how the uncertainty is crucial for the decision maker.

The remainder of the paper is organized as follows.  Section~\ref{s:scenario} enumerates various scenarios that require uncertainty-aware systems, and Section~\ref{s:challenges} details the desired attributes for uncertainty-awareness and surveys our current progress and future work to achieve uncertainty-awareness at both the ML and AI layers. Section~\ref{s:tech_scenario} illustrates how our current results combined with explanations of uncertainty are important to the decision maker. Finally, concluding remarks are given in Section~\ref{s:con}.


\section{Motivational Scenario}
\label{s:scenario}

Our focus is on scenarios where there is a significant open world element, i.e., characterised by prevalence of `unknown unknowns' and `unknown knowns'. This kind of scenario is common in government and public sector domains. For example:
\begin{itemize}
\item In \textit{securing a large-scale event}, consider a case where a known group is observed (e.g., identified from a photo on social media or via a image processing of a CCTV feed), but that group (or type of group) has never been seen before at that particular kind of event. There is no known `political' link between the group's ideology and the event in question making their intent unpredictable using existing predictive models.
\item In \textit{narcotics policing}, consider the example of a drug user being discovered under the affects of a known substance but in an unexpected context, e.g., a drug that is in common urban usage associated with (predictable) gang activity, now discovered in use in a rural location. Again, we have key unknown knowns that make prediction and choice of remedial action challenging.
\item In \textit{public health}, consider an outbreak of a known disease (detectable by established predictive services) in an entirely unexpected location where this disease has never been seen or studied before, and where its spread cannot be reliably predicted, e.g., a tropical disease in a polar context.
\end{itemize}

To focus subsequent discussion on a single concrete example, we will consider the case of route planning in a city that has recently been taken under control by a coalition for peace-keeping. Routes might be unsafe depending on the sentiment of local groups. One of the routes goes through a market, and we have a camera there that can detect for violent activities. Furthermore, the sentiment of the local groups affects their attendance at particular public civic centers, which is directly observable by the coalition. While a Bayesian network model can probabilistically connect these observables to the safety of the various routes, there is very little historical data to learn the parameters of the network or to train a neural network to classify specific local activities at the market. This leads to uncertainty, and Section~\ref{s:tech_scenario} details how uncertainty-aware AI\&ML can help the decision maker avoid or prepare for surprises in this scenario.



\section{Technical Challenge and Solutions}
\label{s:challenges}

In the era of big data, many assume that the parameters of an AI\&ML system can be interpreted as point estimates and the prediction accuracy of the system can generalize as conditions change.  However, for the scenarios described above, training data is sparse and the situations are highly dynamic.  As a result, the parameters are highly uncertain and possibly changing over time. This uncertainty propagates through the process to infer the beliefs in the state values of the latent variables crucial for decision making. When the conditions that the AI\&ML system operates under become vastly different than the training conditions, these uncertainties can be high indicating the inferencing power of the system is currently poor. This section reviews current and suggests future work to understand how to extract these uncertainties.

\subsection{Introduction to Subjective Logic}

Subjective logic is a formalism to understand reasoning under uncertain probabilistic information~\cite{Josang2016-SL}.  It expands the notion of a probability value to a distribution of possible probabilities.  This paper considers variables such as $X$ that can take on values $X=x$ such that $x$ is in the domain $\mathbb{X}$ of cardinality $K = | \mathbb{X}|$. The value of $X$ does change over different instantiations, and there is an underlying ground truth value for the probability $p_{X}(x)$ of $X=x$ for all $x \in \mathbb{X}$.

A subjective opinion can be formed by directly observing $N_{ins}$ independent instantiations of $X$. If over these instantiations, $X=x$ for $n_x$ and assuming an uninformative uniform prior, then the posterior knowledge of the ground truth outcome probability of $X$ is known to follow the Dirichlet distribution
\begin{equation}
\label{eq:Dirichlet distribution}
f_{\bm{\beta}}(\bm{p}_X|\omega_X)= \frac{1}{\beta(\bm{\alpha}_X)}\prod_{x \in \mathbb{X}} p_X(x)^{\alpha_x-1}\;
\end{equation}
for $0 \le p_X(x) \le 1$, where $\beta(\cdot)$ is the $K$-dimensional beta function and the Dirichlet parameters $\alpha_x = n_x+1$ for $x \in \mathbb{X}$ are one particular representation of the opinion $\omega_X$. The opinion $\omega_X$ in belief space is a tuple of beliefs $b_x = \frac{n_x}{s_X}$ for $x \in \mathbb{X}$ and uncertainty $u_X = \frac{K}{s_X}$, where $s_X = \sum_{x \in \mathbb{X}} \alpha_x$ is the Dirichlet strength.


It is also convenient to represent the subjective opinion $\omega_X$ by the mean and Dirichlet strength of the corresponding Dirichlet distribution.  The mean represents the projected probability that converts the opinion into the pignistic probabilities, and is given by
\begin{equation}
P_X(x) = \frac{\alpha_x}{s_X}  \hspace{.1in} \mbox{for $x \in \mathbb{X}$}.
\end{equation}
The variances for the probabilities $p_X(x)$, given by
\begin{equation}
\label{e:pred_var}
\sigma_X^2(x) = \frac{P_X(x)(1-P_X(x))}{s_X+1},
\end{equation}
are functions of the projected probabilities and Dirichlet strength of the subjective opinion.

Subjective logic provides a set of rules to infer the effective Dirichlet strength and projected probabilities of latent variables from the logical entailment with variables for which opinions come from direct observations. For instance, subjective opinions naturally extend to subjective conditional opinions, where for example, the opinion for $X$ conditioned on $Y$ and $Z$ is interpreted as the set $\{\omega_{X|y,z}: y \in \mathbb{Y}, z \in \mathbb{Z} \}$, and $\omega_{X|y,z}$ represents the effective number of times that $X=x$ for $x \in \mathbb{X}$ when $Y=y$ and $Z=z$ while jointly observing $X$, $Y$, and $Z$. Overall, the efficacy of subjective logic reasoning can be evaluated via simulations to determine 1)~how accurate the projected probabilities of the inferred opinions match the ground truth probabilities, and 2)~how accurately the uncertainty (or Dirichlet strength) of the inferred opinions represents the `spread' between the projected and ground truth probabilities.

\subsection{Subjective Logic Bayesian Network}

The Subjective Bayesian network (SBN) was first proposed in \cite{ivanovska.15}; it is an uncertain Bayesian network where the conditionals are subjective opinions instead of point probabilities. In other words, the conditional probabilities are known within a Dirichlet distribution. A SBN reflects the knowledge about a Bayesian network when limited historical data is used to learn the conditionals. The inference in SBN leads to an opinion about the marginal probability of all the unobserved variables conditioned on the values of the observed variables. While different types of SBNs were discussed in \cite{ivanovska.15}, the focus here is on the type that uses the Diriclet distribution interpretation of the subjective opinion to compute uncertainty.  This section reviews subjective belief propagation (SBP) which works for singly-connected networks and is described in detail in \cite{kaplan.18.ijar} for SBNs with binary variables. We are currently extending the inference engine to accommodate general variables whose domain cardinality $K>2$.

SBP extends the Belief Propagation (BP) inference method of Pearl \cite{pearl.86}.  In BP, $\pi$- and $\lambda$-messages are passed from parents and children, respectively, to a node, i.e., variable.  The node uses these messages to formulate the inferred marginal probability of the corresponding variable.  The node also uses these messages to determine the $\pi$- and $\lambda$-messages to send to its children and parents, respectively.  In SBP, the $\pi$- and $\lambda$-messages are subjective opinions characterized by a projected probability and Dirichlet strength.

The SBP formulation approximates output messages as beta-distributed random variables using the methods of moments and a first-order Taylor series approximation to determine the mean and variance of the output messages in light of the beta-distributed input messages. The details of the derivations are provided in \cite{kaplan.18.ijar}.  Given a node $X$ with $m$ parents $U_i$ for $i=1,\ldots,m$, the subjective opinions of the $\pi$-messages sent to $X$ are characterized by the projected probabilities $\pi_{U_i,X}(x)$ and Dirichlet strengths $s_{\pi_{U_i,X}}$.  Likewise, given that $X$ has $k$ children $Y_j$ for $j=1,\ldots,k$, the subjective opinions of the $\lambda$-messages sent to $X$ are characterised by the projected probabilities $\lambda_{U_i,X}(x)$ and Dirichlet strengths $s_{\lambda_{U_i,X}}$.

In the end, SBP computes opinions for each latent variable $X$ conditioned on the state values for all the observed variables. In \cite{kaplan.18.ijar}, it is shown that the projected probabilities of the inferred opinions accurately reflect the probabilities that could be inferred if the ground truth conditional probabilities were known. More importantly for uncertainty-awareness, the uncertainty of the inferred opinions faithfully characterized the accuracy of the projected probabilities. Specifically, empirical results demonstrated that the divergence between the desired and actual confidence bound significance is low. Current and future work is expanding the inference to efficiently work for general (non-binary) variables under general (possibly loopy) network structures.

\subsection{Uncertainty-Aware Neural Network}

\begin{figure}
\begin{center}
\includegraphics[width=3.0in]{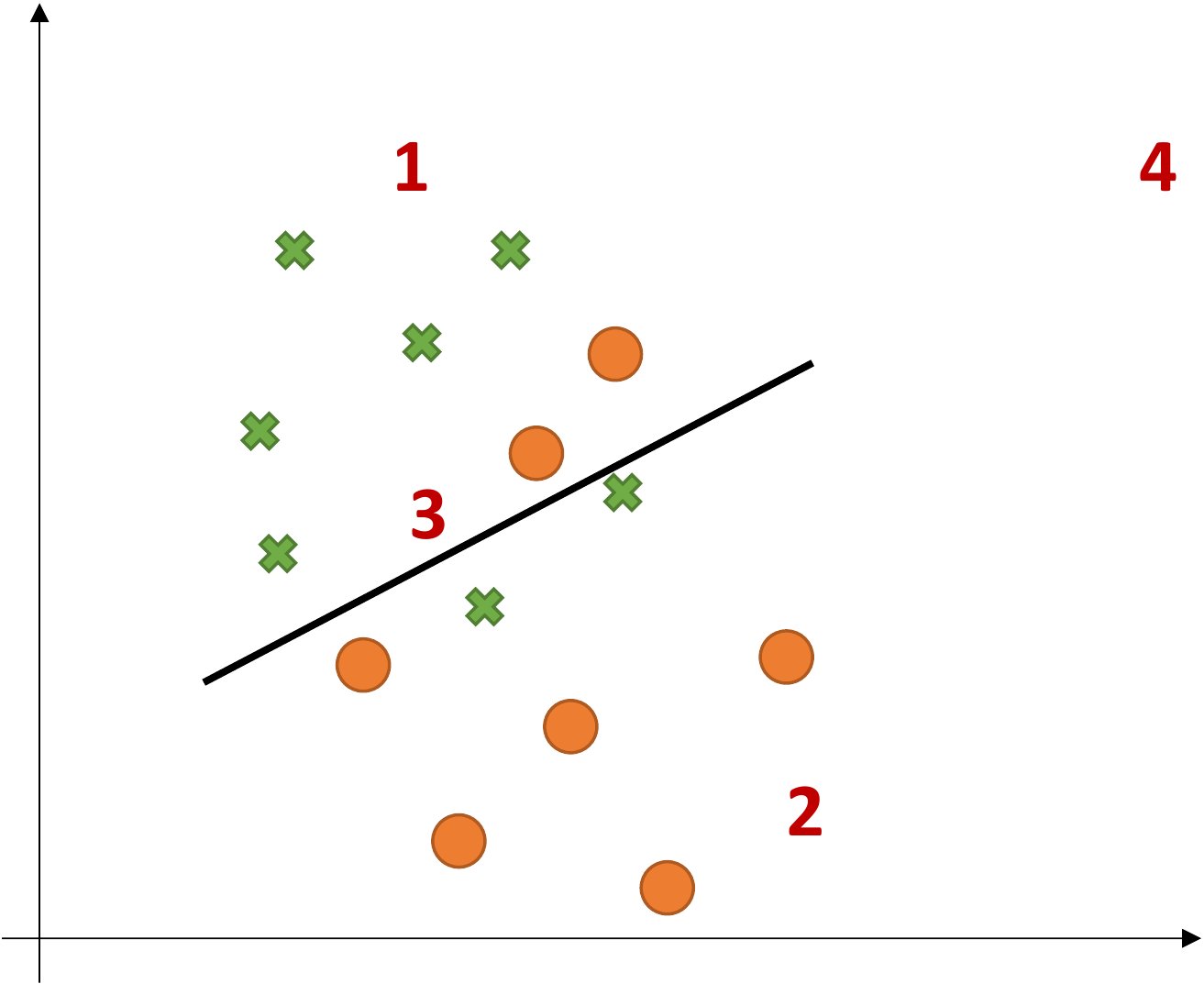}
\caption{Hypothetical feature space of training data for two classes and four test data samples.}
\label{f:sl_class}
\end{center}
\end{figure}

Uncertainty-aware machine learning methods should express confidence based upon the amount of training data of each class that is similar to the data sample under test.  Figure~\ref{f:sl_class} illustrates this concept by visualizing the training and testing data in a feature space. In this example, test samples `1' and `2' should have high beliefs in the `x' and `o' classes, respectively, along with low uncertainty as they are nearby many training samples of the same class. On the other hand, test sample `3' should have beliefs around 0.5 for both the `x' and `o' classes along with low uncertainty as it lies near an equal and large number of training data from the two classes. In other words, the sample lies near a decision boundary. Finally, test sample `4' is far from any of the training data and should have small beliefs for the two classes along with high uncertainty.  Overall, the output opinion from an uncertainty-aware machine learner should represent the amount of nearby training data of the various classes.

It is a challenge to efficiently train a machine learning method to output subjective opinions.  Part of the challenge is to understand how to tabulate distances in the feature space of the training data into class evidences. Furthermore, one must determine a proper feature space, which is most likely not the raw data space, that reduces the effects of noise, invariants, and irrelevant dimensions. Finally, the machine learner must regress the subjective opinion from its model parameters without explicitly storing all of the training data.

Evidential Deep Learning (EDL) aims to extend classical deep learning with the ideas from evidential reasoning and subjective logic~\cite{dempster2008generalization,Josang2016-SL} to quantify uncertainty in classification tasks~\cite{Sensoy2018Evi}.
As mentioned before, subjective logic uses Dirichlet distributions to represent subjective opinions, which encodes belief and uncertainty.
For a sample, EDL tries to learn parameters of a predictive posterior as a Dirichlet density function for the classification of the sample.
This is done by replacing the \textit{softmax} layer in a classical deep classifier with an activation function that produces only non-negative outputs (e.g., \textit{relu}, \textit{softplus} etc.). The resulting output $\bm{e}_X=[e_1,\ldots,e_n]$ is considered as the evidence for the classification of the sample over $n$ class labels. From the calculated evidence, parameters of the corresponding Dirichlet density is calculated as $\bm{\alpha}_X = \bm{e}_X + \bm{1}$.
Hence, when the total evidence is zero, the resulting Dirichlet density corresponds to the uniform distribution, i.e., a Dirichlet density whose all parameters are one. It is mapped to a subjective opinion with zero belief and total uncertainty.

The Dirichlet distribution can be considered as generative process for the categorical distribution over possible class labels, i.e., $\bm{p_X}$. Hence, we can calculate the expected prediction failure for the sample using the following:
\begin{align}
\int ||\bm{y}_X - \bm{p}_X||_2^2 \frac{1}{\beta(\bm{\alpha}_X)}\prod_{x \in \mathbb{X}} p_X(x)^{\alpha_x-1} d \bm{p}_X
\end{align}
where $\bm{y}_{X}$ is the one-hot representation of the class label (i.e., ground truth) and $||\bm{y}_{X} - \bm{p}_X||_2^2$ is the sum square error between the truth and the categorical distribution $\bm{p}_X$.
Hence, the equation calculates the expected sum square error over possible values of $\bm{p}_X$ using the predicted Dirichlet density function.

The loss function consists of the expected sum square error regularized by a term that measures the similarity of the Dirichlet distribution minus the ground truth evidence to the uniform distribution. The regularization term helps to reduce confidence, i.e., increase uncertianty, in cases of conflict. These cases occur in regions of sparse training data, and the regularization term consitions EDL to recognize such cases.

In a set of experiments performed over well-known benchmarks, EDL significantly outperforms state-of-the-art Bayesian Neural Network approaches for uncertainty estimation.
Furthermore, it is more robust to adversarial attacks, since it generates less evidence and increases uncertainty when encountered with adversarial examples.




\subsection{Future Extensions}


Many other AI\&ML approaches can be extended to accommodate uncertainty-awareness beyond Bayesian networks and neural networks. There is a need to incorporate first-order logics into the higher layer reasoning, and it is natural to think about a Markov logic network (MLN)~\cite{richardson.06} where the rule weights are learned within a distribution. However, MLNs are computationally inefficient and perhaps other probabilistic logic reasoning systems such as probabilistic soft logic~\cite{bach2017hinge} or ProbLog~\cite{deraedt:ijcai07} should be considered first.  In fact, collective subjective logic has been proposed as uncertainty-aware extension of probabilistic soft logic~\cite{chen2017csl}. Beyond other reasoning approaches, it also important to understand how to isolate the observations contributing to the uncertainty for each decision variable and how to disseminate the reasoning chain from such observations to the decision variables to the decision maker.

In the near future, we plan to investigate the uncertainty-aware extension of ProbLog.
ProbLog~\cite{deraedt:ijcai07,fierens:tplp15}\footnote{More information on ProbLog, including an open source implementation and an interactive online tutorial, can be found at \url{https://dtai.cs.kuleuven.be/problog/}.} belongs to a family of probabilistic logic programming (PLP) languages~\cite{deraedt:mlj15} following Sato's distribution semantics~\cite{sato:iclp95}.
It extends logic programming by annotating some ground facts with their probability of being true, which generalizes a single program into a distribution over programs that share their rules, but differ in their databases. More specifically, a ProbLog program consists of two parts, a set~$F$ of ground probabilistic facts \verb|p::f| where \verb|p| is a probability and \verb|f| a ground atom, and a set~$R$ of rules \verb|h :- b1,...,bn| where \verb|h| is a logical atom and the \verb|bi| are literals.
\footnote{For the semantics of ProbLog to be well-defined, the set of rules has to have a two-valued well-founded model for each subset of the probabilistic facts: a sufficient condition for this is for programs to be stratified, i.e., have no loops through negation. See \cite{deraedt:mlj15,fierens:tplp15} for further details.}
While the semantics is defined for countably infinite sets of probabilistic facts, see \cite{sato:iclp95} for details, we restrict the discussion to the finite case in the following. ProbLog considers the ground probabilistic facts as independent random variables, i.e., we obtain the following probability distribution $P_F$
over truth value assignments to  sets
of ground facts $F'\subseteq F$:
\begin{equation}
\label{eq:p_f}
P_F(F') = \prod_{f_i\in F'}p_i\cdot\prod_{f_i\in F\setminus F'}(1-p_i)
\end{equation}

As each logic program obtained by choosing a truth value for every
probabilistic fact has a unique least Herbrand model, $P_F$ can be
used to define the \emph{success probability} $P(q)$ of a query $q$, that is,
the probability that $q$ is true in a randomly chosen such program,
as the sum over all programs that entail $q$:
\begin{align}
\label{eq:p_suc}
P(q) &\colonequals \sum\limits_{\substack{F'\subseteq F \\
    \exists\theta F'\cup R\models q\theta}} P_F(F')\\
 &= \sum\limits_{\substack{F'\subseteq F \\
    \exists\theta F'\cup R\models q\theta}} \prod_{f_i\in F'}p_i\cdot\prod_{f_i\in F\setminus F'}(1-p_i)\;.\label{eq:p_suc_facts}
\end{align}

Inference in ProbLog is concerned with computing marginal probabilities of queries, i.e., ground atoms, under this distribution, potentially conditioned on a conjunction of evidence atoms. While this is a \#P-hard problem in general, ProbLog relies on state-of-the-art knowledge compilation techniques to achieve scalable inference across a wide range of models.

The parameters of ProbLog programs can be learned from partial interpretations~\cite{fierens:tplp15}, and ProbLog rules defining a target predicate can be learned from a ProbLog program specifying background knowledge (in the form of facts and/or known rules for other predicates) and ground atoms using the target predicate annotated with target probabilities~\cite{deraedt:ijcai15}.

Inspired by subjective logic, we plan to learn the parameters of the ProbLog programs such as the rule probabilities as beta distributions.  Furthermore, we plan to update the inference engine to accommodate these distribution to output subjective opinions about success probability of the query. This will enable a systematic study of the strengths and weakness of the subjective ProbLog against subjective Bayesian networks.

\section{Theoretical Demonstration}
\label{s:tech_scenario}

To demonstrate the need for uncertainty-aware AI\&ML, we consider a route planning operation within a purposely simplified peace keeping scenario to highlight the salient value of uncertainty-awareness.
In this scenario, the GoodGuys are trying to stabilize a city.  Consider that a commander of the GoodGuys needs to transport some medical supplies from headquarters to a local hospital.  There are three routes through a city populated by two factions, the Capulets and the Montagues.  These factions have tumultuous relationships with each other and the GoodGuys.  One route (Route~A) goes through the center of the Capulet territory, and another route (Route~C) goes through the center of the Montague territory. The remaining route (Route~B) skirts the border of Capulet and Montague territory and passes by the central marketplace.

\begin{figure}[t]
\begin{center}
\includegraphics[width=0.68\linewidth]{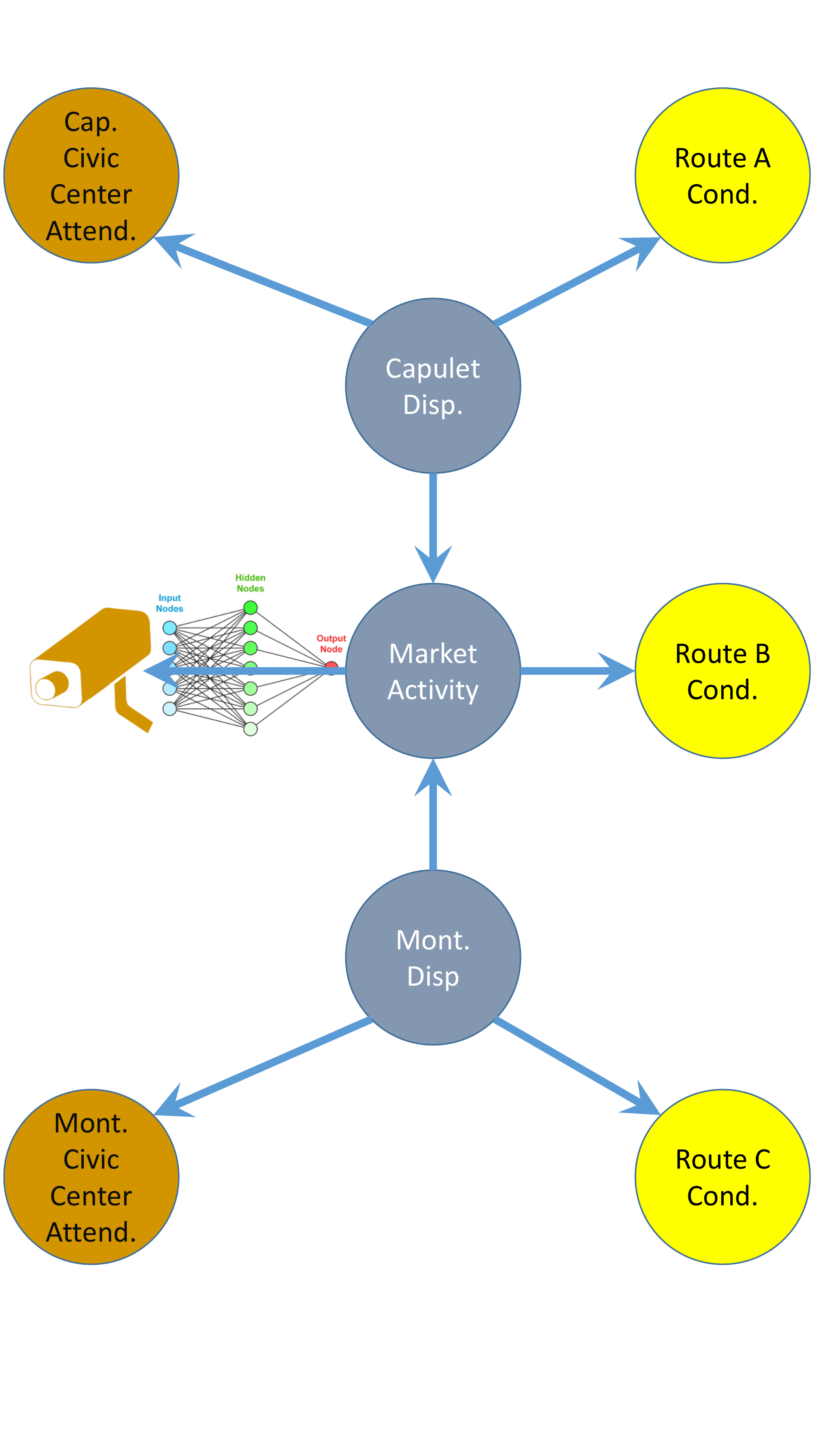}
\end{center}
\caption{Bayesian network model describing the probabilistic interactions of the various variables that influences how to infer route safety in light of the observations including the civic center engagement and marketplace activity likelihood via the surveillance camera.}
\label{f:bayesnet}
\end{figure}

There is a surveillance camera at the marketplace that is connected to a neural network that classifies between normal and violent activity.  The commander also has information about the number of people who visit the community centers every day within the Capulet and Montague territories. Given these observations, the commander wants to decide what route to use and the level of armored escorts needed to protect the transport operation. The commander is using the Bayesian network model illustrated in Figure~\ref{f:bayesnet} to predict the route conditions. This model indicates that the daily dispositions of each faction towards the GoodGuys influences the attendance at the civic centers, the marketplace activity, and the safety of the three routes.

The actual conditional probabilities of the Bayesian network are not known to the GoodGuys.  Instead, they lean a subjective Bayesian network using human intelligence to obtain 100 daily instantiations of the dispositions of the two factions, along with reports of route conditions and civic center attendance. The ground truth conditional probabilities for the Bayesian network is given in Table~\ref{t:condprob}.\footnote{In this example, the variables are binary simply because the current subjective belief propagation method works only for the binary case.  We are currently extending the propagation method to accommodate more general multinomial variables.} The table indicates that the Capulets usually have a favorable disposition towards the Goodguys, in contrast to the Montagues. Whenever either faction has a negative disposition, attendance at the corresponding civic center is higher than normal.  It is also more likely the faction will disrupt any Goodguys traveling through their territory.  Violent activities in the marketplace are only likely to occur when both factions have negative dispositions, and only then is it likely the factions would disrupt GoodGuys traveling along Route~B.  The commander’s knowledge of all these trends is based on the Subjective Bayesian network obtained from the 100 instantiations.

\begin{table*}[t]
\begin{center}
\begin{tabular}{|c|c|c|}\hline
Variable & States & Conditional Probabilities \\ \hline
\begin{tabular}{c} Capulet \\ Disposition \end{tabular} & $\{ {\tt pos}, {\tt neg} \}$ & $P(CD = {\tt pos}) = 0.9$ \\ \hline
\begin{tabular}{c} Montague \\ Disposition \end{tabular} & $\{ {\tt pos}, {\tt neg} \}$ &  $P(MD = {\tt pos}) = 0.1$ \\ \hline
\multirow{4}{*}{\begin{tabular}{c} Civic Center\\ Attendance \end{tabular}} & \multirow{4}{*}{$\{ {\tt norm}, {\tt high} \}$} & $P(CCA = {\tt norm}|CD = {\tt pos}) = 0.8$ \\
&&$P(CCA = {\tt norm}|CD = {\tt neg}) = 0.1$ \\ \cline{3-3}
&&$P(MCA = {\tt norm}|MD = {\tt pos}) = 0.8$ \\
&&$P(MCA = {\tt norm}|MD = {\tt neg}) = 0.1$ \\ \hline
\begin{tabular}{c} Marketplace \\ Activity \end{tabular} &$\{ {\tt norm}, {\tt violent} \}$ & \begin{tabular}{c} $P(MA = {\tt norm}|CD = {\tt pos}\vee MD = {\tt pos}) = 0.99$ \\
$P(MA = {\tt norm}|CD = {\tt neg}\wedge MD = {\tt neg}) = 0.01$ \end{tabular}\\ \hline
\multirow{4}{*}{\begin{tabular}{c} Route\\Condition \end{tabular} } & \multirow{4}{*}{$\{ {\tt safe}, {\tt danger} \}$} & $P(RA = {\tt safe}| CD = {\tt pos}) = 0.9$\\
&& $P(RA = {\tt safe}| CD = {\tt neg}) = 0.1$ \\ \cline{3-3}
&& $P(RB = {\tt safe}| MA = {\tt norm}) = 0.9$ \\
&& $P(RB = {\tt safe}| MA = {\tt violent}) = 0.1$ \\ \cline{3-3}
&& $P(RC = {\tt safe}| MD = {\tt  pos}) = 0.9$ \\
&& $P(RC = {\tt safe}| MD = {\tt neg}) = 0.1$ \\ \hline
\end{tabular}
\caption{Ground truth conditional probabilities for the Bayesian network in Figure~\ref{f:bayesnet}.}
\label{t:condprob}
\end{center}
\end{table*}

Table~\ref{t:sbn_infer} provides the inferred opinions of the route conditions from the subjective Bayesian network for each of the three routes for various observations. In the absence of any observations, the uncertainty of all there opinions is relatively low and reflective of 100 effective pieces of evidence. The opinions for routes~A and~B are comparable with a safety and danger belief of about 0.78 and 0.20, respectively, and these beliefs are flipped for the Route~C opinion.  These opinions are consistent with general trends from the ground truth conditional probabilities.

\begin{table*}[t]
\begin{center}
\begin{tabular}{|c|ccc|ccc|ccc|} \hline
 & \multicolumn{3}{|c|}{Route~A} & \multicolumn{3}{|c|}{Route~B} & \multicolumn{3}{|c|}{Route~C}\\
Observations & $b_{\tt safe}$ & $b_{\tt dang}$ & $u$ & $b_{\tt safe}$ & $b_{\tt dang}$ & $u$ & $b_{\tt safe}$ & $b_{\tt dang}$ & $u$ \\ \hline
none & 0.78 & 0.20 & 0.02 & 0.77 & 0.21 & 0.02 & 0.21 & 0.77 & 0.02 \\ \hline
\begin{tabular}{c} $CCA$ = {\tt norm} \\ $MCA$ = {\tt high} \\ $\omega_{MA}^{\mbox{cam}} = (0.95,0,0.05)$\end{tabular} & 0.92 & 0.06 & 0.02& 0.91 & 0.07 & 0.02& 0.13 & 0.84 & 0.03 \\ \hline
\begin{tabular}{c} $CCA$ = {\tt norm} \\ $MCA$ = {\tt norm} \\ $\omega_{MA}^{\mbox{cam}} = (0.95,0,0.05)$\end{tabular} & 0.90 & 0.08 & 0.02 & 0.91 & 0.07 & 0.02 & 0.54 & 0.37 & {\bf 0.09} \\ \hline
\begin{tabular}{c} $CCA$ = {\tt norm} \\ $MCA$ = {\tt norm} \\ $\omega_{MA}^{\mbox{cam}} = (0,0.95,0.05)$\end{tabular} & 0.66 & 0.15 & {\bf 0.19} & 0.02 & 0.50 & {\bf 0.48} & 0.54 & 0.37 & {\bf 0.09} \\ \hline
\begin{tabular}{c} $CCA$ = {\tt norm} \\ $MCA$ = {\tt norm} \\ $\omega_{MA}^{\mbox{cam}} = (0,0,{\bf 1})$\end{tabular} & 0.88 & 0.09 &  0.03 & 0.79 & 0.08 & {\bf 0.13} & 0.55 & 0.36 & {\bf 0.09}\\ \hline
\end{tabular}
\caption{Opinions for the route conditions obtained via subjective belief propagation in light of various observations.  Note that camera likelihood opinion $\omega_{MA}^{\mbox{cam}} = (b_{\tt norm},b_{\tt violence},u)$ is the backpropagated message to the marketplace activity variable from the camera data.}
\label{t:sbn_infer}
\end{center}
\end{table*}

When the camera is detecting normal activity at the marketplace, and the attendance at the civic centers in Capulet and Montague sections are normal and high, respectively, the route conditions opinions demonstrate low uncertainty.  The uncertainty does increase slightly over the previous case of no observation because the observations do constrain the number of effective training instantiations that are appropriate for the inference. Since these observations are likely to occur, the uncertainty remains relatively small, and observations increases the beliefs that routes~A and~B are safe and Route~C is dangerous. With these observations, the commander can choose Route~A or~B with limited armor support.

When the attendance at the Montague civic centers goes normal, then uncertainty about Route~C increase while the opinions of routes-A and~B are unaffected.  This is due to the fact that it is unlikely for the Montagues to have a positive disposition towards the GoodGuys, which reduces the amount of effective training samples to determine the condition or Route~C.  Again the commander can choose Route~A or~B with limited armor support.

Then when the camera has high confidence of violence at the marketplace, the uncertainty for the opinions of routes~A and-B increases drastically. This is due the fact that it is extremely unlikely for violence to occur in the marketplace when attendance at both civic centers are at normal levels. There are very few training instantiations to help determine these opinions. On the other hand, the opinion for Route~C has not changed as it still reflects the oddity of normal attendance at the Montague civic center. In this case, the commander would need to pick either Route~A or Route~C where the probability of safety of Route~A is higher but at more than twice the uncertainty. In either case the commander may want heavier armor support.

Finally, the neural network reports that it is uncertain about the activity class. This primary affects the uncertainty about Route~B.  Furthermore, resolution of this activity classification has tremendous influence over the subjective opinions of routes~A and~B as can be seen in the third and fourth rows of Table~\ref{t:sbn_infer}.  Therefore, the AI\&ML system alerts the commander that it is uncertain due to its processing of the surveillance camera data.  The commander directs his team to analyze the video footage. The team determines that the people in the marketplace have synchronized in a dance routine that the neural network was never trained to recognize. The commander digs further with some experts to learn that the dance is actually an ancient ritual that two tribes in the region perform to synchronize the mind and body when the tribes are preparing to attack a common exogenous enemy. In light of this additional context, the commander decides to provide heavy armed support on both the ground and air to escort the supply truck.

The scenario illustrates how understanding uncertainty is important for the commander to assess risk.  The uncertainty can arise because the observations are due to rare events for which sparse training means few exemplars to establish opinions for the states of the decision variables. The uncertainty can also arise when the neural networks cannot interpret the input data.  This case can represent a `black swan’ event that the neural network was never trained to understand.  The synchronized dance is an example of an unknown unknown that requires human investigation.

\section{Conclusions}
\label{s:con}

AI\&ML systems have shown great promise as decision making aids. However, such systems are not impervious to making mistakes. Many such mistakes occur because the systems are operating in situations that they were not trained to understand.  Instead of forcing a decision, it is important for such a system to be able to characterize its uncertainty relative to the similarity of the current observations to the training data, and to explain the uncertainty to a human decision maker.  This way, the human can trust the AI\&ML system to crunch though the large piles of data at a rate faster than the human and alert the human when the system realizes that it no longer can provide quality inferences. By isolating the uncertainty, the system can then explain the source and nature of the uncertainty to the user.  This enables the human to focus his/her reasoning strengths to the special cases that need extra attention.  This paper highlights some recent work in subjective Bayesian network and evidential neural networks to eventually achieve this uncertainty-awareness.  Some other future directions and challenges are discussed.  We believe that the development of uncertainty-aware AI\&ML systems will become an important avenue of research for the community.

\section*{Acknowledgement}
This research was sponsored by the U.S. Army Research Laboratory and the U.K. Ministry of Defence under Agreement Number W911NF-16-3-0001. The views and conclusions contained in this document are those of the authors and should not be interpreted as representing the official policies, either expressed or implied, of the U.S. Army Research Laboratory, the U.S. Government, the U.K. Ministry of Defence or the U.K. Government. The U.S. and U.K. Governments are authorized to reproduce and distribute reprints for Government purposes notwithstanding any copyright notation hereon.

\bibliographystyle{aaai}
\bibliography{biblio.bib}

\end{document}